\documentclass{article}
\usepackage{ijcai23}

\usepackage{times}
\usepackage{soul}
\usepackage{url}
\usepackage[hidelinks]{hyperref}
\usepackage[utf8]{inputenc}
\usepackage[small]{caption}
\usepackage{graphicx}
\usepackage{amsmath}
\usepackage{amsthm}
\usepackage{booktabs}
\usepackage{algorithm}
\usepackage{algorithmic}
\usepackage[switch]{lineno}

\usepackage{multirow}
\DeclareMathOperator*{\argmin}{arg\,min}
\usepackage{amsfonts}
\usepackage{amsthm}
\usepackage{booktabs}
\usepackage{float}
\usepackage{algorithm}
\usepackage{subfig}
\usepackage{algorithmic}
\usepackage{tikz}
\usepackage[edges]{forest}
\usepackage{multirow}
\usepackage{multicol}
\usepackage{tabularx}
\usepackage{soul}
\usepackage{eqnarray}
\usepackage{makecell}
\usepackage{comment}
\usepackage{array}
\usepackage{longtable}
\usepackage{amssymb}% http://ctan.org/pkg/amssymb
\usepackage{pifont}% http://ctan.org/pkg/pifont

\usepackage{makecell}
\usepackage{comment}
\usetikzlibrary{trees,positioning,shapes,shadows,arrows}
\newenvironment{linez}
 {\par\nopagebreak
  \parindent0pt
  }
 {\par
  \nopagebreak
  \noindent\ignorespacesafterend}

\usetikzlibrary{trees,positioning,shapes,shadows,arrows}

\title{A Survey on Dataset Distillation: Approaches, Applications and Future Directions}

\iftrue
\author{
${\textbf{Jiahui Geng}}^1$\and
${\textbf{Zongxiong Chen}^*}^2$\and
${\textbf{Yuandou Wang}}^3$\and
${\textbf{Herbert Woisetschläger}}^4$\and \\
${\textbf{Sonja Schimmler}}^2$\and 
${\textbf{Ruben Mayer}}^{4}$ \and
${\textbf{Zhiming Zhao}}^{3}$\And  
${\textbf{Chunming Rong}}^{1}$  \\
\affiliations
$^1$University of Stavanger \\
$^2$ Fraunhofer FOKUS\\
$^3$ University of Amsterdam\\
$^4$ Technical University of Munich
\emails
\{jiahui.geng, chunming.rong\}@uis.no,
\{herbert.woisetschlaeger, ruben.mayer\}@tum.de,\\
\{z.zhao, y.wang8\}@uva.nl,
\{zongxiong.chen, sonja.schimmler\}@fokus.fraunhofer.de
}
\fi

\begin{document}

\maketitle
\def\thefootnote{\textbf{*}}\footnotetext{Corresponding author: zongxiong.chen@fokus.fraunhofer.de}

\begin{abstract}
Dataset distillation is attracting more attention in machine learning as training sets continue to grow and the cost of training state-of-the-art models becomes increasingly high. By synthesizing datasets with high information density, dataset distillation offers a range of potential applications, including support for continual learning, neural architecture search, and privacy protection. Despite recent advances, we lack a holistic understanding of the approaches and applications. Our survey aims to bridge this gap by first proposing a taxonomy of dataset distillation, characterizing existing approaches, and then systematically reviewing the data modalities, and related applications. In addition, we summarize the challenges and discuss future directions for this field of research.
\end{abstract}

\section{Introduction}
High-quality and large-scale datasets are crucial for the success of deep learning, not only enabling the development of end-to-end learning systems~\cite{schmidhuber2015deep,nmtBahdanau}, but also serving as benchmarks to evaluate different machine learning architectures~\cite{deng2009imagenet,koehn-2005-europarl}. However, the explosion of deep learning dataset sizes has posed considerable challenges concerning processing, storage, and transfer. Training neural networks often require thousands of iterations on the entire dataset, which consumes significant computational resources and power. Tasks such as hyperparameter optimization~\cite{maclaurin2015gradient} and neural architecture search (NAS)~\cite{such2020generative} are even more resource-intensive. One promising solution is to use smaller datasets with high information density to reduce resource consumption while preserving model performance.

% Recent studies have shown that it is possible to use a subset of the original data to train neural networks, resulting in models with competitive performance. Approaches such as curriculum learning~\cite{graves2017automated}, active learning~\cite{konyushkova2017learning}, and coreset selection~\cite{sener2017active} have demonstrated this. 
Research in the area of curriculum learning~\cite{graves2017automated}, active learning~\cite{konyushkova2017learning}, and coreset selection~\cite{sener2017active} has shown that it is possible to sample a subset of the original data to train neural networks, resulting in models with competitive performance. This also implies that we can train high-performance models with less effort while downstream tasks like continual learning (CL)~\cite{castro2018end,prabhu2020gdumb}, neural architecture search (NAS) will also benefit. Nevertheless, creating an algorithm-agnostic, efficient, and unbiased small dataset to replace the original is still challenging. For instance, coreset selection is typically an NP-hard problem, making it computationally intractable and difficult to apply to large datasets. 

An alternative approach to coreset is dataset distillation, which aims to distill the original data onto a smaller synthetic dataset~\cite{wang2018dataset}. Dataset distillation techniques have continued to evolve, with various methods such as gradient matching~\cite{zhao2021datasetdc}, trajectory matching~\cite{cazenavette2022dataset}, and kernel ridge regression~\cite{nguyen2020dataset} being proposed to optimize the distilled data, resulting in improved distillation performance in terms of both the accuracy of the trained model on the test set and the generalization capability across different network architectures. However, there remain challenges regarding optimization stability and computational efficiency.

Despite the recent advancements in dataset distillation, a comprehensive overview summarizing its advances and applications is currently not available. This paper aims to fill this gap by presenting a taxonomy of dataset distillation. To our knowledge, it is the first work that provides a systematic categorization of the different methods and techniques used in dataset distillation. The paper mainly makes the following contributions:
\begin{itemize}
\item We propose a novel taxonomy of dataset distillation, which can help researchers to better understand the research landscape and find their areas of interest. 
\item We present existing distillation approaches in detail, discussing their strengths and weaknesses; 
\item We discuss important challenges in this domain, highlighting promising directions for future research. 
\end{itemize}

%     \item We first propose a novel taxonomy of dataset distillation. We introduce the principles, advantages and disadvantages of different methods, and summarize the general techniques.

%     \item We summarize the applications in different data modalities, e.g. image, text, audio and graph data. In addition, we summarize the experimental settings and evaluation metrics, which will help researchers design experiments to evaluate innovative algorithms.

%     \item We conclude by highlighting some open problems and promising directions for future research.
% \end{itemize}

% \subsection{Outline}
The paper is organized as follows. In Section~\ref{sec:taxo}, we first present our taxonomy of dataset distillation. Then, we introduce the learning frameworks and common enhancement methods in Section~\ref{sec:framework} and Section~\ref{sec:enhancement}, respectively. Section~\ref{sec:modality} summarizes the advances in different data modalities. In Section~\ref{sec:application}, we categorize the related applications according to the dataset distillation properties. Finally, we conclude this paper with future directions in Section~\ref{sec:conclusion}. 

\begin{figure}[t]
% \vspace{-3mm}
    \centering
    % \hspace{-0.2cm}%
    % \includegraphics[width=0.5\textwidth]{img/taxo.png}
    % \includegraphics[width=0.5\textwidth]{img/taxo-3.pdf}
    % \includegraphics[width=0.48\textwidth]{img/taxo-Page-2.drawio-1.pdf}
    \includegraphics[width=0.48\textwidth]{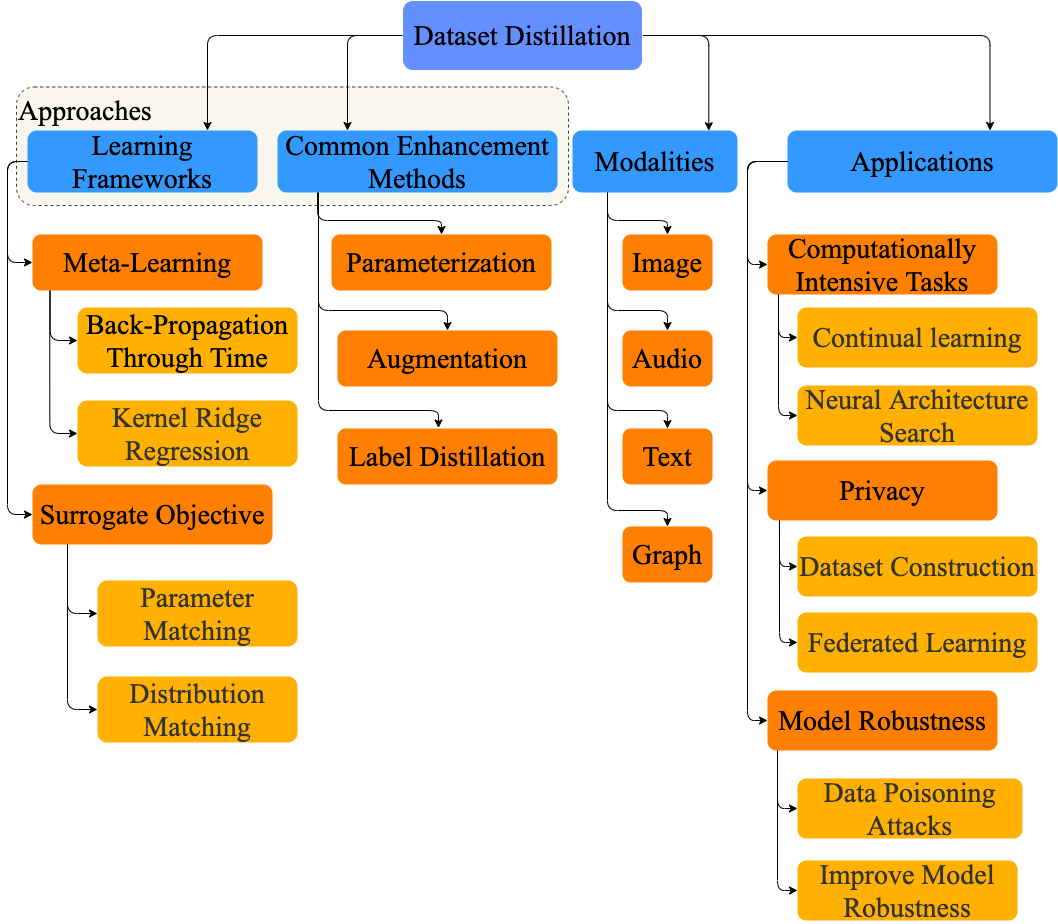}
    % \vspace{-2mm}
    \caption{Taxonomy of dataset distillation.} 
    \label{fig:taxonomy}
% \vspace{-3mm}
\end{figure}

\section{Taxonomy}
\label{sec:taxo}
\subsection{Basics of Dataset Distillation}
We begin by introducing the key notations used in this paper. $\mathcal{D}$ represents a general dataset, $f_{\theta}$ represents a neural network with parameters $\theta$, and $f_{\theta}(x)$ denotes the model's prediction for data point $x$. The expected loss for dataset $\mathcal{D}$ in relation to $\theta$ is defined as

\begin{small}
\begin{equation}
    \mathcal{L}_{\mathcal{D}}(\theta) = \mathbb{E}_{(x,y) \sim P_{\mathcal{D}}} [\ell (f_{\theta}(x),y)],
\end{equation}
% \vspace{-3mm}
\end{small}
\begin{linez}
where $x$ and $y$ are the input data and label pair from $\mathcal{D}$, $\ell (f_{\theta}(x),y)$ is the given loss value between the prediction and ground truth.
\end{linez}

Dataset distillation aims to reduce the size of large-scale training input and label pairs  $ \mathcal{T} = \{(x_i, y_i) \}_{i=1}^{\lvert \mathcal{T}\rvert} $ by creating smaller synthetic pairs $\mathcal{S} = \{(\hat{x}_j, \hat{y}_j)\}_{j=1}^{\lvert \mathcal{S} \rvert}$, so that models trained on both $\mathcal{T}$ and $\mathcal{S}$ can achieve similar performance, which can be formulated as:
\begin{small}
\begin{equation}\label{eq:loss-pairing}
    \mathcal{L}_{\mathcal{T}}(\theta^{\mathcal{S}}) \simeq \mathcal{L}_{\mathcal{T}}(\theta^{\mathcal{T}}),
\end{equation}
\end{small}
\begin{linez}
where $\theta^{\mathcal{S}}$ and $\theta^{\mathcal{T}}$ are the parameters of the models trained on $\mathcal{S}$ and $\mathcal{T}$ respectively.
\end{linez}

% \scriptsize{Hello world}
% \zongxiong{TODO: figure size not fit in page}

% \begin{figure}[H]
% % \vspace{-5mm}
%     \centering
%     \hspace{-0.6cm}%
%     \includegraphics[width=0.5\textwidth]{img/taxo-2.pdf}
%     % \includegraphics[width=0.5\textwidth]{img/taxo-3.pdf}
%     \vspace{-2mm}
%     \caption{Taxonomy of dataset distillation.} 
%     \label{fig:taxonomy}
% \vspace{-5mm}
% \end{figure}

\subsection{Taxonomy Explanation}
The taxonomy of dataset distillation is illustrated in Figure~\ref{fig:taxonomy}. In this taxonomy, we classify the research about dataset distillation from three perspectives: approaches, data modalities and applications. The approaches can be decomposed into two parts. In the learning framework, we explain how dataset distillation can be modeled, optimized and solved in different ways, such as using meta-learning~\cite{andrychowicz2016learning} or surrogate objectives (see Section~\ref{sec:surrogate}). Meta-learning can be further divided into using back-propagation through time and using kernel ridge regression. Surrogate objective can be subdivided into parameter matching and distribution matching. We categorize the common enhancement methods, which can be plugged into a learning framework, mainly into parameterization (see Section~\ref{sec:parameterization}), augmentation (see Section~\ref{sec:aug}) and label distillation (see Section~\ref{sec:labeldistillation}). 
% According to data modality, existing work can be classified into four types of data: image, audio, text and graph. According to the goals, we classify applications into computationally intensive tasks, including continual learning and neural architecture search; privacy protection, including dataset construction and federated learning; and model robustness, including implementing data poisoning attacks and improving model robustness. This hierarchical 
Existing work can be classified into four types of data: image, audio, text, and graph, based on data modality. Applications can be further divided into three categories: computationally intensive tasks such as continual learning and neural architecture search, privacy protection including dataset construction and federated learning, and model robustness, encompassing data poisoning attacks and improving robustness. 
% This taxonomy offers a clear overview of the research landscape.
Corresponding to our taxonomy, some representative papers, together with their characteristics, have been listed in Table~\ref{tab-comp}. It comprehensively compares learning frameworks, enhancement methods, data modality, and applications.  
% Specially, we list the learning framework, enhancement methods, modalities of the experimented datasets, and applications. 

\renewcommand\arraystretch{1.3}   % 2表示2倍行高。一般用1.1~1.3会比较好看。

\begin{table*}[h!]
\begin{center}

\setlength{\tabcolsep}{1.3mm}{
\begin{tabular}{ m{3.6cm} | m{2.0cm} | p{1.9cm} | m{0.3cm}|m{0.3cm}|m{0.3cm}|m{0.3cm} | p{2.6cm} | p{1.8cm} | p{1.5cm} }
% \begin{tabularx}{\textwidth}{ c |X | c | c|c|c|c | c | c | c }
% \begin{tabular}{ m{3.5cm} m{1.6cm} m{0.3cm} m{0.3cm} m{0.3cm} m{0.3cm}  p{1.9cm}  p{7cm} }
% \begin{longtable}{ m{3.5cm} | m{1.6cm} | m{0.3cm}|m{0.3cm}|m{0.3cm}|m{0.3cm} | p{1.9cm} | p{7cm} }
   
    % \toprule
    % \hline
    \Xhline{1.5pt}
    % \hline
    % \textbf{Author} & \textbf{Objective} & \textbf{PM} & \textbf{DM} & \textbf{IM} \\ \hline
    
    \multirow{3}{*}{\textbf{Paper}} & 
    \multicolumn{1}{l|}{
        \multirow{2}{*}{
        \textbf{\noindent Learning}}
        } & 
    % \multicolumn{1}{*}{\textbf{Enhancement}} & 
    \multicolumn{1}{l|}{
        \multirow{2}{*}{
        \textbf{Enhancement}}
    } & 
    \multicolumn{4}{c|}{
        \multirow{1}{*}{\textbf{Data Modality}} 
        } & 
    \multicolumn{3}{c}{
        \multirow{1}{*}{\textbf{Applications}}
    }
    \\\cline{4-10}
    & \textbf{Framework} & \textbf{Methods} & 
    \rotatebox[origin=c]{270}{\textbf{Image}} & \rotatebox[origin=c]{270}{\textbf{Text}} & \rotatebox[origin=c]{270}{\textbf{Audio}} & \rotatebox[origin=c]{270}{\textbf{Graph}} & 
    \textbf{Computationally Intensive Tasks} & \textbf{Privacy} & \textbf{Robustness}

    \\
    % \hline
    % \midrule
    \Xhline{1.0pt}
    
%%% Meta Learning
    DD~\cite{wang2018dataset} & \multirow{6}{1pt}{Back-Propagation Through Time}  &  & \checkmark & &  &
    & & & Data Poisoning
    % First to propose the concept of dataset distillation 
    \\\cline{0-0}\cline{3-10}

    SLDD, TDD~\cite{sucholutsky2021soft} &   & LD
    & \checkmark & \checkmark & & &  & &
    % First experimented with text dataset and increased performance by introducing learnable distilled labels 
    \\\cline{0-0}\cline{3-10}
    Addressable Memory~\cite{deng2022remember} &  & Factorization, LD & \checkmark & & & 
    & CL & & 
    %Suggested to distill a dataset into addressable memory for memory saving and effiective query 
    \\\hline
    KIP~\cite{nguyen2020dataset,nguyen2021dataset} & \multirow{3}{1pt}{Kernel Ridge Regression} & LD & \checkmark  &  &  &
    & & $\rho$-corruption &
    % First to use kernel-ridge regression with close-form solution to avoid bi-level optimization 
    \\\cline{0-0}\cline{3-10}
    
    FRePo~\cite{zhou2022dataset} &  &  LD & \checkmark & & & 
    & CL & MIA & 
    % Proposed neural feature regression with pooling to overcome overfitting and computational complexity 
    \\\cline{0-0}\cline{3-10}
    
    RFAD~\cite{loo2022efficient} &  & LD & \checkmark &  & &
    & & $\rho$-corruption &
    % Suggested to use random feature approaximation of the NNGP to reduce computational complexity    
    \\\hline
%%% Gradient Matching
    DC~\cite{zhao2021datasetdc} &  \multirow{10}{1pt}{Parameter Matching} &   & \checkmark  & & &
    & CL, NAS & & 
    % First work to use surrogate objective for training
    \\\cline{0-0}\cline{3-10}
    
    DSA~\cite{zhao2021datasetdsa} &  & DSA & \checkmark & & &  
    & CL, NAS & & 
    % Designed a efficient data augmentation 
    \\\cline{0-0}\cline{3-10}
    
    MTT~\cite{cazenavette2022dataset} & & DSA & \checkmark & & & 
    & & & 
    % Proposed to improve the performance with long-range parameter matching 
    \\\cline{0-0}\cline{3-10}
    
    IDC~\cite{kim2022dataset} & &  Factorization, DSA& \checkmark & & \checkmark & 
    & CL & & 
    % Proposed a multi-formation process to create multiple synthetic data and demonstrated the first work in speech domain 
    \\\cline{0-0}\cline{3-10}

    HaBa~\cite{liu2022dataset} & & Factorization & \checkmark &  & \checkmark &  &
    CL & & 
    % Proposed a hallucinator-basis factorization using to hallucinator encode inner relations between samples 
    \\\cline{0-0}\cline{3-10}

    PSG~\cite{chen2022private} & &  & \checkmark & & &  &
     & MIA, DP & 
    % Proposed a hallucinator-basis factorization using to hallucinator encode inner relations between samples 
    \\\cline{0-0}\cline{3-10}
    
    GCond~\cite{jin2021graph} &  &  &   & & &  \checkmark
    & NAS & & 
    % First work to distill discrete graph dataset 
    \\\cline{0-0}\cline{3-10}

    DosCond~\cite{jin2022condensing} &  &   & & &  & \checkmark
    & & & 
    % Bernoulli distribution and used one-step gradient matching to boost the computational efficiency
    \\\hline
    
%%%% Distribution Matching
    DM~\cite{Zhao2021DatasetCW} & \multirow{7}{1pt}{Distribution Matching} & DSA & \checkmark & & & 
    & CL, NAS & & 
    % First to propose distribution matching to avoid bi-level optimization and enable parallel computation 
    \\\cline{0-0}\cline{3-10}

    CAFE~\cite{wang2022cafe} &  &   & \checkmark & & & 
    & & & 
    % Further developed the distribution matching with feature alignment and other methods  
    \\\cline{0-0}\cline{3-10}

    IT-GAN~\cite{zhao2022synthesizing} & & DSA, GAN & \checkmark & & & 
    & & & 
    %Investigated a schema to convert a pre-trained GAN into a targeted data generator  
    \\\cline{0-0}\cline{3-10}

    GCDM~\cite{liu2022graph} & &   & & & & \checkmark & 
     & & 
    % Optimized the graph distillation through the distributino matching                             
    \\\cline{0-0}\cline{3-10}

    KFS~\cite{lee2022dataset} & &   & \checkmark & & & 
    & & & 
    % Proposed to efficiently factorize the data generating process into latent codes and decoders
    \\ \hline
%%% Kernel 

    % \hline
    \Xhline{1.5pt}
\end{tabular}}
% \end{longtable}
\end{center}
\vspace{-3mm}
\caption{Summary of existing dataset distillation works. CL -- Continual Learning, NAS -- Neural Architecture Search, MIA -- Membership Inference Attack, DP -- Differential Privacy, and LD -- Label Distillation. Note: \checkmark -- if it uses such data modality.}
\label{tab-comp}
\vspace{-3mm}
\end{table*}

\section{Learning Frameworks}\label{sec:framework}
According to the learning goals, the current learning frameworks can mainly be divided into two categories: meta-learning methods based on inner model performance and methods using surrogate objectives.
% Dataset distillation frameworks are divided into two categories. One is to optimize synthetic data based on performance trained on synthetic data, which corresponds to meta-learning. 
% The other uses a surrogate objective, making the newly learned model close to the model trained on the original model in the parameter or feature distribution space.
\subsection{Meta-Learning}\label{sec:metalearning}
Meta-learning~\cite{andrychowicz2016learning} refers to learning about learning, and often refers to machine learning algorithms that learn from the output of other machine learning algorithms. In this problem, the distilled data are treated as hyperparameters and the objective is to optimize the distilled data in a bi-level optimization problem as follows:
\begin{small}
\begin{equation}
\label{eq:meta}
      \mathcal{S^*} = \argmin_{\mathcal{S}} \mathcal{L}_{\mathcal{T}} (\theta^{\mathcal{S}}) 
\:\: \text{s.t.}  \:\: \theta^{\mathcal{S}}   = \argmin_{\mathcal{\theta}} \mathcal{L}_{\mathcal{S}} (\theta),   
\end{equation}
\end{small}
\begin{linez}
where the inner loop, optimizing $\theta^{\mathcal{S}}$, trains a model on the synthetic dataset until convergence, and the outer loop, optimizing $\mathcal{S}$, subsequently optimizes the synthetic dataset, so that the model has good generalization capability and can perform well on the real dataset. The distillated dataset is optimized using the meta-gradient: 
\end{linez}
\begin{small}
\begin{equation}
    \mathcal{S} \leftarrow \mathcal{S} - \alpha \nabla_{\mathcal{S}} \mathcal{L}_{\mathcal{T}}(\theta^\mathcal{S}),
\end{equation}
\end{small}
\begin{linez}
where $\alpha$ is the learning rate for updating the synthetic dataset. 
\end{linez}
\begin{figure}[t]
% \vspace{5mm}
    % \hspace{-0.6cm}%
%    \includegraphics[width=0.51\textwidth]{img/DD-bilevel-2.pdf}
    \includegraphics[width=0.5\textwidth]{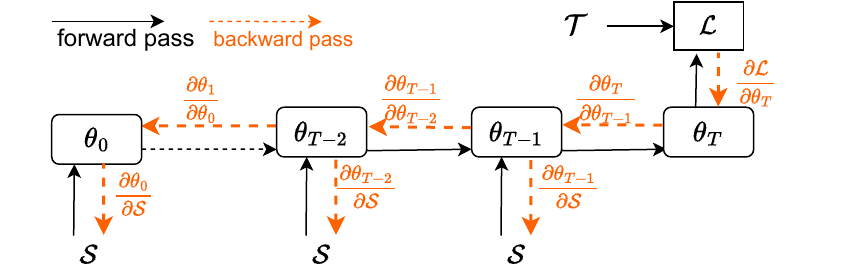}

    % \vspace{-2mm}
    \caption{Back-Propagation Through Time. The gradient $\nabla_\mathcal{S} \mathcal{L}$ is calculated via back-propagation through time (see orange dashed line). 
    % In the figure, we omits the details about meta-gradients corresponding to optimizer, i.e., $\eta$, $m$, etc. DD~\protect\cite{wang2018dataset} and SLDD~\protect\cite{sucholutsky2021soft} are the cases when $T=1$, whereas in AddMem~\protect\cite{deng2022remember}, $T$ reaches up to 200.
    } 
    \label{fig:bptt}
% \vspace{-5mm}
\end{figure}

\subsubsection{Back-Propagation Through Time}
Computing the meta-gradient $\nabla_{\mathcal{S}} \mathcal{L}_{\mathcal{T}}(\theta^\mathcal{S})$ requires differentiating through inner optimization. When the model is learned in an iterative way, i.e., 
\begin{small}
\begin{equation}\label{eq:multi-bptt-t}
    \theta_{t+1} = \theta_{t} - \eta \nabla_{\theta_t} \ell(f_{\theta}(\hat{x}), \hat{y}),
\end{equation}
\end{small}
\begin{linez}
where $\eta$ is the learning rate for inner loop and meta-gradient is calculated by back-propagation through time (BPTT):
\end{linez}
\begin{small}
 \begin{equation}
 \label{eq:bptt}
     \nabla_{\mathcal{S}} \mathcal{L}_{\mathcal{T}}(\theta^\mathcal{S}) = \frac{\partial \mathcal{L}}{\partial \mathcal{S}} = \frac{\partial \mathcal{L}}{\partial \theta_T} \Big( \sum\limits_{t=0}^{t=T} \frac{\partial \theta_T}{\partial \theta_t} \cdot \frac{\partial \theta_t}{\partial \mathcal{S}} \Big)
 \end{equation}
\end{small}
\begin{linez}
which is illustrated in Figure~\ref{fig:bptt}. It is evident that the computation overhead is high due to the recursive calculation of the meta-gradient using Equation~\ref{eq:bptt}.
\end{linez}

% Obviously, recursively computation of the meta-gradient using Equation~\ref{eq:bptt} causes high computation overhead. 

To make the implementation of Equation~\ref{eq:bptt} feasible, ~\cite{wang2018dataset} suggest using the Truncated Back-Propagation Through Time (TBPTT) method, which involves unrolling the inner-loop optimization steps as a single step of gradient descent optimization,
\begin{small}
\begin{equation}\label{eq:dd}
    \hat{x}, \hat{\eta} = \argmin\limits_{\hat{x}, \hat{\eta}} \ell(f_{\theta_1}(x), y), \: \text{s.t.} \:\: \theta_1 = \theta_0 - \eta \nabla_{\theta_0} \ell(f_{\theta_0}(\hat{x}), \hat{y}),
\end{equation}
\end{small}
\begin{linez}
where $\hat{x}$, $\hat{y}$ are synthetic dataset and $\hat{\eta}$ the learning rate for the optimizer. 
% We see that Equation~\ref{eq:dd} is the realization of Equation~\ref{eq:multi-bptt-t} when doing one-step gradient descent.. 
\end{linez}

\cite{deng2022remember} further improves the learning framework by incorporating a momentum term and extending the length of unrolled trajectories. Empirical results show that the momentum term can consistently improve performance and that longer unrolled trajectories can lead to better model parameters that produce more efficient gradients for compressed representation learning.

BPTT methods have been criticized for several issues, as noted in~\cite{zhou2022dataset}: 1) high computational cost and memory overhead; 2) bias in short unrolls; 3) gradients exploding or vanishing in long unrolls; and 4) chaotic conditioned loss landscapes.

\subsubsection{Kernel Ridge Regression}\label{sec:krr}
\begin{figure}[h]
% \vspace{5mm}
    \centering
    \includegraphics[width=0.42\textwidth]{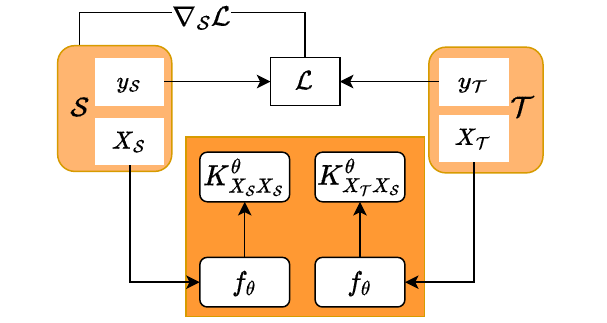}
    % \vspace{0mm}
    \caption{Kernel Ridge Regression. The figure shows the workflow of kernel ridge regression. The details refer to Equation~\ref{eq:kernel-ridge-regression} and \ref{eq:frepo}. 
    The key difference is that KIP~\protect\cite{nguyen2020dataset} uses NTK kernel, RFAD~\protect\cite{loo2022efficient} uses Neural Network Gaussian Process (NNGP)  kernel. Feature extractor $f_\theta$ in FRePo~\protect\cite{zhou2022dataset} is parameterized during training.}    
    % Another key difference between FRePo and KIP, RFAD is that }
    \label{fig:dd-krr}
% \vspace{-4mm}
\end{figure}

\cite{nguyen2020dataset} transform dataset distillation into a kernel ridge regression (KRR) problem, where the synthetic set is used as the support set and the original set as the target set. Their approach result in a closed-form solution in terms of convex optimization, simplifying the expensive nested optimization into first-order optimization (see Figure~\ref{fig:dd-krr}). They introduce the Kernel-Inducing Point (KIP) algorithm which utilizes neural tangent kernel (NTK)~\cite{jacot2018neural} ridge regression to compute the exact outputs of an infinite-width neural network trained on the synthetic set, bypassing the need for gradient and back-propagation computation on any neural network. The KRR loss function for a given kernel and batch data from synthetic set $(X_{\mathcal{S}}, y_{\mathcal{S}})$ evaluated on batch data from real set $(X_{\mathcal{T}}, y_{\mathcal{T}})$ can be formulated as,
\begin{comment}
\begin{small}
\begin{equation}
    \mathcal{L}_{\mathcal{T}}(\mathcal{S}) = \frac{1}{2} \lVert y_{\mathcal{T}} - K_{X_{\mathcal{T}}X_{\mathcal{S}}}(K_{X_{\mathcal{S}}X_{\mathcal{S}}} + \lambda I)^{-1} y_{\mathcal{S}} \rVert^2,
\label{eq:kernel-ridge-regression}
\end{equation}
\end{small}
\end{comment}
\begin{small}
\begin{equation}
    \argmin_{X_\mathcal{S}, y_\mathcal{S}} \frac{1}{2} \lVert y_{\mathcal{T}} - K_{X_{\mathcal{T}}X_{\mathcal{S}}}(K_{X_{\mathcal{S}}X_{\mathcal{S}}} + \lambda I)^{-1} y_{\mathcal{S}} \rVert^2,
\label{eq:kernel-ridge-regression}
\end{equation}
\end{small}
\begin{linez}
where $K_{X_{\mathcal{T}}X_{\mathcal{S}}}$ is the Gram matrix of $X_{\mathcal{S}}$ and $X_{\mathcal{T}}$, and $K_{X_{\mathcal{S}}X_{\mathcal{S}}}$ is the Gram matrix of $X_{\mathcal{S}}$.
\end{linez}

\cite{zhou2022dataset} propose a novel method, neural feature regression with pooling (FRePo), which utilizes a more flexible conjugate kernel with neural features to replace the NTK in KIP~\cite{nguyen2020dataset}. This approach breaks down the traditional KRR training pipeline into two components: a feature broke $f_\theta$ and a linear classifier. When calculating the meta-gradient of $\mathcal{S}$, FRePo fixes the feature extractor parameters and updates $\mathcal{S}$ $T$ times according to Equation~\ref{eq:kernel-ridge-regression}, where $T$ is a hyperparameter that helps prevent the support/synthetic dataset from memorizing a specific network. Additionally, a model pool is employed to alleviate overfitting in the distillation process.
\begin{small}
\begin{align}\label{eq:frepo}
    K^{\theta}_{X_{\mathcal{T}}X_{\mathcal{S}}} &= f_{\theta}(X_{\mathcal{T}}) f_{\theta}(X_{\mathcal{S}})^{\top}, \\
    K^{\theta}_{X_{\mathcal{S}}X_{\mathcal{S}}} &= f_{\theta}(X_{\mathcal{S}}) f_{\theta}(X_{\mathcal{S}})^{\top}
\end{align}
\end{small}

\cite{loo2022efficient} propose to use random feature approximation for distillation (RFAD), which utilizes random feature approximation of the Neural Network Gaussian Process (NNGP) kernel to replace the NTK used in KIP. This approach reduces the computation of the Gram matrix to $\mathcal{O}(\lvert \mathcal{S}\rvert)$, which is linear with the size of the synthetic set, compared to $\mathcal{O}(\lvert \mathcal{S}\rvert^2)$, the complexity of accurately calculating the NTK kernel matrix. They also suggest using cross-entropy loss with Platt scaling~\cite{platt1999probabilistic} to provide a more accurate probabilistic interpretation for classification tasks.

% In KIP~\protect\cite{nguyen2020dataset}, they uses a NTK kernel, whereas in RFAD~\protect\cite{loo2022efficient}, they use Neural Network Gaussian Process (NNGP) kernel instead. FRePo~\protect\cite{zhou2022dataset} employ networks as neural kernel functions.

\subsection{Surrogate Objective}\label{sec:surrogate}
Instead of optimizing directly based on model performance, surrogate objective approaches optimize a proxy objective, such as the parameters or gradients of the model. These approaches assert that the effectiveness of a model trained on a full dataset and a distilled dataset can be inferred from their corresponding parameters and gradients.

\subsubsection{Parameter Matching}\label{sec:parameter}
 In contrast to optimizing directly based on the loss value corresponding to the distilled data, it aims to make the model approximate the original model in the parameter space, i.e. $\theta^{\mathcal{S}} \approx \theta^{\mathcal{T}}$. 
Empirically, the trajectory of parameters vary with its initial state $\theta_0$. Therefore, the objective of parameter matching should be agnostic to the initialization.
When distance between the model parameters trained on the synthetic dataset and the real dataset are consistently close, the distilled dataset can be considered as a good alternative of original whole dataset. Let $\theta^{\mathcal{S}}(\theta_0)$, $\theta^{\mathcal{T}}(\theta_0)$ denote the trained models from the same initialization $\theta_0$, the objective function can be expressed as:
\begin{small}
\begin{equation}
\label{eq:pm}
   \min_{\mathcal{S}} \mathbb{E}_{\theta_0 \sim P_{\theta_0}} [D(\theta^{\mathcal{S}}(\theta_0), \theta^{\mathcal{T}}(\theta_0))],   
\end{equation}
\end{small}
\begin{linez}
where $D(\cdot, \cdot)$ is a distance function. 
\end{linez}
% Here, we distinguish the difference between $D$ and $\mathcal{L}$ 

To enable a more guided optimization and apply the incomplete training, DC~\cite{zhao2021datasetdc} synthesizes images by minimizing the gradient matching loss at each training step $t$:
\begin{small}
\begin{equation}\label{eq:gm}
    \min_{\mathcal{S}} \mathbb{E}_{\theta_0 \sim P_{\theta_0}} [ \sum_{t=0}^{T-1} D(\nabla_{\theta} \mathcal{L}_{\mathcal{S}}(\theta_t), \nabla_{\theta} \mathcal{L}_{\mathcal{T}}(\theta_t)) ]
\end{equation}
\end{small}
where $T$ is the hyperparameter for the number of training iterations. 

\cite{cazenavette2022dataset} suggest overcoming bias accumulated from one-step gradient by matching training trajectories (MTT). MTT considers the training trajectories ${\theta}_{t=0}^{T-1}$ on real data as the expert models, the model $\hat{\theta}$ trained on the synthetic dataset as the student model.
It randomly samples $\theta_t^\mathcal{T}$ from the expert model to initialize the student model, and the objective is to make the student model $\hat{\theta}^{\mathcal{S}}_{t+N}$ approximate the expert model $\theta^{\mathcal{T}}_{t+M}$ after $N$ iterations. The optimization objective is given by
\begin{comment}
\begin{small}
\begin{equation}
    \mathcal{L} = \frac{\lVert \hat{\theta}_{t+N} - \theta_{t+M} \rVert_2^2}{\lVert \theta_t -\theta_{t+M}\rVert_2^2},
\end{equation}
\end{small}
\end{comment}
\begin{small}
\begin{equation}
    D = \frac{\lVert \hat{\theta}^{\mathcal{S}}_{t+N} - \theta^{\mathcal{T}}_{t+M} \rVert_2^2}{\lVert \theta^{\mathcal{T}}_t -\theta^{\mathcal{T}}_{t+M}\rVert_2^2},
\end{equation}
\end{small}
\begin{linez}
where $M, N$ are the hyperparameters.
\end{linez}
% \zongxiong{@Geng, I changed $\mathcal{L}$ to $D$ to make symbol consistent in this section}

Parameter matching methods are often criticized for: 1) high bias it introduces~\cite{wang2022cafe}. The synthetic set learned by gradient matching is extremely biased towards those large gradient samples, which will decrease its generalization capability on unseen architectures; 2) expensive bi-level optimization. For example training 50 images/class using DC~\cite{zhao2021datasetdc} requires 500K epochs of updating network parameter $\theta_t$ and 50K updating of $\mathcal{S}$; and 3) fragile hyper-parameters~\cite{Zhao2021DatasetCW} tuning. e.g. how often to update $\theta_t$ and $\mathcal{S}$ in DC, as well as $M, N$ in MTT~\cite{cazenavette2022dataset} are critical.

\begin{figure*}[h!]
% \vspace{-5mm}
    \centering
    % \hspace{-1.5cm}\subfloat[Parameter Matching]% {\includegraphics[trim={0 0 0 0},clip,width=0.53\textwidth]{img/DD-gm-2.pdf}\label{fig:dd-gm-sub}}
    % \hspace{8mm}
    % \hspace{2cm}\subfloat[]{\includegraphics[width=1\textwidth]{img/DD-gm-dm.pdf}\label{fig:dd-dm-sub}}
    \includegraphics[width=1\textwidth]{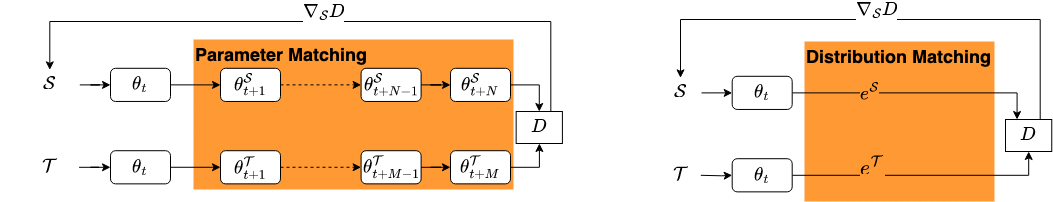}\label{fig:dd-dm-sub}
    % \vspace{-5mm}
    \caption{Surrogate Objective. The figure presents the workflow of parameter matching (left) and distribution matching (right). The key difference between algorithm DC~\protect\cite{zhao2021datasetdc} and MTT~\protect\cite{cazenavette2022dataset} is that DC uses information from one-step optimization (gradient) while MTT using parameters after several steps. Definition of $D$ is given as in Equation~\ref{eq:pm}.
    % $\mathcal{S}$ corresponds to released synthetic dataset, and by applying specific processing functions, e.g. augmentation (see Sec.~\ref{sec:aug}), 
    % normalization (see Sec.~\ref{sec:norm}),
    % GAN (see Sec.~\ref{sec:gan}), factorization (see Sec.~\ref{sec:factor}), etc, on it can obtain $\mathcal{S'}$. 
    In distribution matching, the embeddings $e^{\mathcal{S}}$ and $e^{\mathcal{T}}$ in DM~\protect\cite{Zhao2021DatasetCW} are extracted from layer output of ConvNet and the ${D}$ is maximum mean discrepancy, whereas, $e^{\mathcal{S}}$ and $e^{\mathcal{T}}$ in CAFE~\protect\cite{wang2022cafe} correspond to layer-wise features and ${D}$ is a mean square error.}
    \label{fig:dd-surrogate}
% \vspace{-5mm}
\end{figure*}
\subsubsection{Distribution Matching}\label{sec:distributionmatching}
The objective of distribution matching is essentially to learn synthetic samples so that the distribution of the synthetic samples is similar to that of real samples in the feature space. They use an empirical estimate of the maximum mean discrepancy (MMD) as a metric to evaluate the distance of spatial distribution. Due to the high computational complexity and difficulty in optimization caused by high dimensionality, Zhao~\cite{Zhao2021DatasetCW} use different randomly initialized neural networks as feature extractors to reduce the input dimension to low-dimensional space. 
\begin{comment}
\begin{small}
\begin{equation}
\label{eq:dm}
\hspace{-2mm}
    \min_{\mathcal{S}} \mathbb{E}_{\theta \sim P_{\theta}\atop \omega \sim \Omega} \bigg\lVert \frac{1}{\lvert \mathcal{S} \rvert} \sum_{i=1}^{\lvert \mathcal{S} \rvert} f_{\theta}(\mathcal{A}(\hat{x}_i, \omega)) - \frac{1}{\lvert \mathcal{T} \rvert} \sum_{i=1}^{\lvert \mathcal{T} \rvert} f_{\theta}(\mathcal{A}(x_i, \omega)) \bigg\rVert^2,
\end{equation}
\end{small}
\end{comment}
\begin{small}
\begin{equation}
\label{eq:dm}
% \hspace{-2mm}
    \min_{\mathcal{S}} \mathbb{E}_{\theta \sim P_{\theta}} \bigg\lVert \frac{1}{\lvert \mathcal{S} \rvert} \sum_{i=1}^{\lvert \mathcal{S} \rvert} f_{\theta}(\hat{x}_i) - \frac{1}{\lvert \mathcal{T} \rvert} \sum_{i=1}^{\lvert \mathcal{T} \rvert} f_{\theta}(x_i) \bigg\rVert^2,
\end{equation}
\end{small}
\begin{linez}
where $f_\theta$ is parameterized by $\theta$, and $\theta$ is sampled from a random distribution $P_\theta$. $\lvert \mathcal{S} \rvert$ and $\lvert \mathcal{T} \rvert$ are the cardinality of dataset $\mathcal{S}$ and $\mathcal{T}$, respectively.
\end{linez}
% $\mathcal{A(\cdot, \omega)}$ is DSA operator ~\cite{zhao2021datasetdsa} (see Section~\ref{sec:aug})

To better capture the whole dataset distribution,~\cite{wang2022cafe} propose to use layer-wise feature alignment in CAFE to learn a more comprehensive characteristic of the distribution. They also introduce a loss function to improve the discriminative ability of the learned samples. The classification loss is calculated using the feature centers of real sample and averaged synthetic samples of each class.

\begin{comment}
\begin{figure}[h]
% \vspace{-2mm}
    \centering
    \includegraphics[width=0.5\textwidth]{img/DD-dm-2.pdf}
    \vspace{-5mm}
    \caption{Distribution Matching. In DM, the embedding $e$ is extracted from one layer of ConvNet and the $\mathcal{L}$ is MMD. Whereas, $e$ in CAFE corresponds to a collection of layer-wise features and $\mathcal{L}$ is MSE with respect to layer-wise features.}
    \label{fig:dd-dm}
\vspace{-3mm}
\end{figure}
\end{comment}

% Inevitably, some information will be lost. 
\section{Common Enhancement Methods}\label{sec:enhancement}
In this section we introduce some techniques that can be integrated into the learning framework presented in the previous section to further enhance distillation performance.
\subsection{Parameterization} \label{sec:parameterization}
% Methods that directly synthesize data do not take into account any data regularity. 
Dataset parameterization aims to utilize the regularity to guide the synthesis, It helps enhance the interpretability by learning hidden patterns, and control the diversity of the synthetic data. In~\cite{zhao2022synthesizing}, the authors propose IT-GAN, a method that uses a pre-trained GAN decoder to increase the informativeness distilled data. IT-GAN first obtains latent vectors from training samples using GAN Inversion~\cite{abdal2019image2stylegan}, then it use the distribution matching algorithm to learn the latent vectors. These vectors can be fed into a pre-trained GAN decoder to induce synthetic images of the original size. In addition, most distillation methods processes each synthetic sample independently, ignoring mutual consistency and relationships between samples. Factorization are proposed to decompose images into different parts to better capture the correlation between different samples and improve the diversity. IDC~\cite{kim2022dataset} utilizes a multi-formation function as the decoder as the decoder to store more information in single sample. ~\cite{deng2022remember} propose to learn matrix-based codes and decodes and use matrix multiplication to generate synthetic datasets. ~\cite{lee2022dataset} employ the latent code - decoder mode for factorization. The decoder is designed as an upsampling neural network containing three ConvTranspose2d layers, aiming to restore latent codes compressed in low dimensions into the image pixel space. ~\cite{liu2022dataset} propose HaBa, which chooses to decompose the image into two parameter spaces of basis and hallucinator. Where hallucinator is an encoder-transformation-decoder structure. Specifically, the encoder is composed of CNN blocks, followed by an affine transformation with scale $\sigma$ and a decoder of a symmetric CNN architecture. 

\subsection{Augmentation}\label{sec:aug}
In~\cite{zhao2021datasetdsa}, the authors propose using differentiable siamese augmentation (DSA) when learning synthetic images, which leads to more informative datasets. DSA is a pluggable technique that includes operators like \textit{scale}, \textit{flip}, \textit{crop}, \textit{rotate}, \textit{color jitters}, and \textit{cutout}. It can be easily integrated into various distillation methods and has been widely used in~\cite{Zhao2021DatasetCW,wang2022cafe}. In~\cite{cui2022dc}, DSA is found to achieve the best performance compared to other data augmentation techniques. However, current augmentation techniques are not suitable for discrete data such as graphs and text.

\subsection{Label Distillation}\label{sec:labeldistillation}
 Label distillation relaxes the restrictions on labels, allowing them to have richer semantics beyond one-hot vectors. It is first introduced in SLDD~\cite{sucholutsky2021soft} and has been shown to improve not only the storage efficiency but also the distillation performance. Their method only requires to make the label in  Equation~\ref{eq:dd} learnable variables.~\cite{nguyen2020dataset} also provide a label learning algorithm based on the closed-form solution in KRR.  It is reported that only five distilled images from MNIST would enable the model to achieve 92$\%$ accuracy~\cite{sucholutsky2021soft}.

 \section{Data Modalities}\label{sec:modality}
\label{sec:modality}
Dataset distillation, first proposed for images, has been applied to various modalities. In this section, we categorize existing works according to data modality and discuss some of the challenges.
% \ref{tab-comp}
\subsection{Image}
Most dataset distillation methods to date have been performed on image datasets~\cite{wang2018dataset,nguyen2020dataset,zhou2022dataset,zhao2021datasetdc,kim2022dataset,cazenavette2022dataset}. These works have constructed benchmarks to facilitate fair comparisons of novel approaches. Images have a continuous real-value domain, which allows direct optimization of synthetic images using deep learning optimizers. We find that experimental datasets become increasingly complex, starting from MNIST, CIFAR10, and SVHN, to more challenging datasets like TinyImageNet and ImageNet~\cite{zhou2022dataset,cazenavette2022dataset,kim2022dataset}. Furthermore, parameterization methods that capture on the regularity of images are becoming increasingly prevalent in the field, as evidenced by recent research such as ~\cite{kim2022dataset,zhao2022synthesizing,liu2022dataset}.

\subsection{Audio}
Speech signals also satisfy the regularity condition of a low-rank data subspace, i.e., temporally adjacent signals haves similar spectra. Therefore, many parametrization methods~\cite{liu2022dataset,kim2022dataset} designed for image dataset can also be applied in this domain. They both experiment with the Speech Commands~\cite{warden2018speech} dataset. In detail, they process the waveform data with a short-time Fourier transform to obtain the magnitude spectrogram and used log-scale magnitude spectrograms for the experiments. Their works show that dataset distillation can achieve consistent performance on downstream tasks of speech signals.

\subsection{Text}
The discrete nature poses challenges to textual distillation. ~\cite{sucholutsky2021soft} first embed the text into a contiguous space using pre-trained GloVe embedding and fill or truncate all sentences to a pre-determined length. In this way, each sentence can be regarded as a single channel image of size length $\times$ embedding dimension.  Text distillation also involves finding the nearest embedding in the dictionary for each vector in the optimized matrix, and transforming these embeddings into the corresponding words and finally the sentence. 

Current efforts are based on primitive bi-level optimization, which is computationally inefficient. There is a lack of work analyzing factors such as the difficulty of the dataset, sentence length, or cross-architecture generalization. Distilled sentences may consist of unrelated words, which makes it difficult to interpret and further analyze. Exploring ways to leverage regularity and context in text distillation is a promising area of research. 

\subsection{Graph}
Graph data is very common in real life, e.g. social networks, Web relationship analysis, and user-item interaction can all be modeled as graph data containing nodes and edges.
\cite{jin2021graph,jin2022condensing} design a strategy to simultaneously compress node features and structural information based on gradient matching. ~\cite{liu2022graph} adopt the distribution matching to boost the performance and show that the dataset distillation was significantly efficient and in some datasets they reached $95\%$ of  the original performance by compressing $99\%$ of the data. Graph distillation is mainly challenged by heterogeneous, abstract, high-level graph representations.

\section{Applications}\label{sec:application}

Dataset distillation, initially designed for model training acceleration, has shown potential in various applications due to its properties.

\subsection{Computationally Intensive Tasks}
\subsubsection{Continual Learning}
Continual learning (CL) addresses the problem of catastrophic forgetting by using strategies such as experience replay, which stores representative samples from previous tasks as a buffer to recall knowledge. Dataset distillation, which involves highly compressed representations, is an alternative to traditional sampling methods. There are currently two experimental settings for using distillation in CL. ~\cite{zhao2021datasetdc,zhao2021datasetdsa} use different datasets, SVHN, MNIST, and USPS, three handwritten digit recognition datasets, and take EEIL~\cite{castro2018end} as the baseline for continuous learning. In the study of~\cite{Zhao2021DatasetCW}, the experimental settings are changed to incremental class learning on the CIFAR100 dataset. The researchers establish a baseline using the GDumb method ~\cite{prabhu2020gdumb} and randomly divided 100 classes into 5 and 10 learning steps, with 20 and 10 classes per step respectively.

\subsubsection{Neural Architecture Search}
Neural architecture search (NAS) is known to be expensive as it involves training multiple architectures on the entire training dataset and selecting the best-performing one on the validation set. To address this issue, researchers have proposed using a distilled dataset as a proxy of the entire dataset, which can effectively identify the best network. Related experiments on the CIFAR10 dataset have been reported in DC~\cite{zhao2021datasetdc}, DSA~\cite{zhao2021datasetdsa}, and DM~\cite{Zhao2021DatasetCW}. These studies construct a search space of 720 ConvNets by varying hyperparameters such as network depth, width, activation function, normalization, and pooling layers over a uniform grid. The effectiveness of the distilled dataset was evaluated using the Spearman's rank correlation coefficient between the validation accuracies obtained by the proxy dataset and the entire dataset. A higher correlation value indicates that the proxy dataset is more reliable.

\subsection{Privacy}
\subsubsection{Dataset Construction}
Machine learning is vulnerable to a variety of privacy attacks, such as membership inference attacks~\cite{shokri2017membership}, model inversion attacks~\cite{fredrikson2015model,carlini2023extracting}, and gradient inversion attacks~\cite{geng2021towards,geng2023improved}, where attackers attempt to infer task-independent private information from the target model, and even recover the original training data. Additionally, data collection and publishing raise privacy and copyright concerns. ~\cite{dong2022privacy,zhou2022dataset} have shown that models trained on synthetic data are robust to both loss-based and likelihood-based membership inference attacks. To ensure that the distilled samples cannot be inferred from real ones, \cite{nguyen2020dataset} implemented the $\text{KIP}_{\rho}$ variant, which randomly initialized $\rho$ proportion of each image and kept them unchanged during training. This idea was later followed by $\text{RFAD}_{\rho}$ \cite{loo2022efficient}. ~\cite{chen2022private} added a differential privacy (DP) mechanism~\cite{dwork2006differential} to the distillation process to provide rigorous privacy guarantees. Medical data often requires strict anonymization before publication, ~\cite{li2022dataset} propose to dataset distillation to construct privacy-preserving datasets.

\subsubsection{Federated Learning}
Federated learning (FL) is an emerging technology that enables different clients to collaboratively train a shared model without sharing their local data. It faces challenges such as high bandwidth requirements for uploading large model updates and a lack of strict privacy guarantees. There are several works that propose to combine dataset distillation in FL. ~\cite{hu2022fedsynth,xiong2022feddm} suggest sharing lightweight synthetic datasets instead of sharing model updates, since the distilled dataset size is generally smaller. However, this may introduce bias and increase the computational load, which can negatively impact the performance and efficiency of FL.

\subsection{Robustness}
\subsubsection{Data Poisoning Attacks}
Distilled data lose its fidelity and may not be visually distinguishable from its original contents, making it vulnerable to data poisoning attacks and difficult to detect. Studies have shown that a small number of these poisoned samples can significantly reduce the accuracy of a model's predictions on a specific category. ~\cite{wang2018dataset} propose a study on data poisoning attacks using dataset distillation.~\cite{liu2023backdoor} propose two backdoor attacks on distilled data by injecting triggers into the synthetic data during the distillation process, either in the initial stage or throughout the entire process.

\subsubsection{Improve Model Robustness}
Dataset distillation can also be used as a means of improving its robustness. Researchers have proposed using optimization techniques to learn a robust distilled dataset, such that a classifier trained on this dataset will have improved resistance to adversarial attacks. ~\cite{tsilivis2022can} have combined the KIP method with adversarial training to enhance the robustness of the distilled dataset. ~\cite{wu2022towards} approached the problem of dataset learning as a tri-level optimization problem to obtain a distilled dataset that minimizes robust error on the data-parameterized classifier.

\section{Conclusion and Future Directions}
\label{sec:conclusion}

In this paper, we present a systematic review of recent advances in dataset distillation. We introduce a novel taxonomy that categorizes existing works from various perspectives. We find that most existing efforts are geared toward image datasets, whereas the handling of discrete text and graph data remains a significant challenge. There is a limited exploration of robustness, and further research is necessary as the technology gains wider adoption. Our study demonstrates the research landscape in this field and suggests directions for future work.

\subsection{Computational Efficiency}
The computational efficiency of dataset distillation is an important consideration, as many current methods for dataset distillation can be computationally expensive, particularly for larger datasets. The goal of dataset distillation is to reduce the size of a dataset while preserving its key features and patterns, but this process often requires complex optimization and clustering algorithms, which can be computationally intensive. Methods like MTT~\cite{cazenavette2022dataset}, KIP~\cite{nguyen2020dataset}, and FRePo~\cite{zhou2022dataset} can cause GPU memory bottlenecks when the number of images per class (IPC) increases. While the DM~\cite{Zhao2021DatasetCW} approach proposes using distribution matching to avoid model training, and RFAD~\cite{loo2022efficient} proposes using NNGP to reduce the computational complexity of kernel ridge regression, the computational efficiency of distillation still requires improvement, particularly for larger datasets. 

\subsection{Performance Degradation on Larger IPC}
According to \cite{cui2022dc}, current dataset distillation methods perform well only when the number of images per class (IPC) is relatively small. As the IPC increases, the performance of most distillation methods deteriorates and becomes similar to that of random sampling. Therefore, it is important to explore whether dataset distillation can overcome this limitation and maintain superior performance on larger datasets.

\subsection{Weak Labels}
Currently, research on dataset distillation primarily focuses on classification tasks. However, its potential for more complex tasks, such as image detection and segmentation, named entity recognition, summarization, and machine translation, remains untapped. Exploring the technique's effectiveness on these tasks could provide deeper insights into data characteristics and the inner workings of AI.

\section*{Acknowledgements}
This work is partially funded by the European Union‘s Horizon 2020 Research and Innovation Program through Marie Skłodowska-Curie Grant 860627 (CLoud ARtificial Intelligence For pathologY (CLARIFY) Project). We also sincerely thank \href{https://github.com/Guang000/Awesome-Dataset-Distillation}{Awesome-Dataset-Distillation} for its comprehensive and timely DD publication summary.

% \section*{Contribution Statement}

\appendix

% \bibliographystyle{named}
% \bibliography{ijcai23}

\bibliographystyle{named}
\bibliography{ijcai23}

\begin{thebibliography}{}

\bibitem[\protect\citeauthoryear{Abdal \bgroup \em et al.\egroup
  }{2019}]{abdal2019image2stylegan}
Rameen Abdal, Yipeng Qin, and Peter Wonka.
\newblock Image2stylegan: How to embed images into the stylegan latent space?
\newblock In {\em Proceedings of the IEEE/CVF International Conference on
  Computer Vision}, pages 4432--4441, 2019.

\bibitem[\protect\citeauthoryear{Andrychowicz \bgroup \em et al.\egroup
  }{2016}]{andrychowicz2016learning}
Marcin Andrychowicz, Misha Denil, Sergio Gomez, Matthew~W Hoffman, David Pfau,
  Tom Schaul, Brendan Shillingford, and Nando De~Freitas.
\newblock Learning to learn by gradient descent by gradient descent.
\newblock {\em Advances in neural information processing systems}, 29, 2016.

\bibitem[\protect\citeauthoryear{Bahdanau \bgroup \em et al.\egroup
  }{2015}]{nmtBahdanau}
Dzmitry Bahdanau, Kyunghyun Cho, and Yoshua Bengio.
\newblock Neural machine translation by jointly learning to align and
  translate.
\newblock In Yoshua Bengio and Yann LeCun, editors, {\em 3rd International
  Conference on Learning Representations, {ICLR} 2015, San Diego, CA, USA, May
  7-9, 2015, Conference Track Proceedings}, 2015.

\bibitem[\protect\citeauthoryear{Carlini \bgroup \em et al.\egroup
  }{2023}]{carlini2023extracting}
Nicholas Carlini, Jamie Hayes, Milad Nasr, Matthew Jagielski, Vikash Sehwag,
  Florian Tram{\`e}r, Borja Balle, Daphne Ippolito, and Eric Wallace.
\newblock Extracting training data from diffusion models.
\newblock {\em arXiv preprint arXiv:2301.13188}, 2023.

\bibitem[\protect\citeauthoryear{Castro \bgroup \em et al.\egroup
  }{2018}]{castro2018end}
Francisco~M Castro, Manuel~J Mar{\'\i}n-Jim{\'e}nez, Nicol{\'a}s Guil, Cordelia
  Schmid, and Karteek Alahari.
\newblock End-to-end incremental learning.
\newblock In {\em Proceedings of the European conference on computer vision
  (ECCV)}, pages 233--248, 2018.

\bibitem[\protect\citeauthoryear{Cazenavette \bgroup \em et al.\egroup
  }{2022}]{cazenavette2022dataset}
George Cazenavette, Tongzhou Wang, Antonio Torralba, Alexei~A Efros, and
  Jun-Yan Zhu.
\newblock Dataset distillation by matching training trajectories.
\newblock In {\em Proceedings of the IEEE/CVF Conference on Computer Vision and
  Pattern Recognition}, pages 4750--4759, 2022.

\bibitem[\protect\citeauthoryear{Chen \bgroup \em et al.\egroup
  }{2022}]{chen2022private}
Dingfan Chen, Raouf Kerkouche, and Mario Fritz.
\newblock Private set generation with discriminative information.
\newblock {\em arXiv preprint arXiv:2211.04446}, 2022.

\bibitem[\protect\citeauthoryear{Cui \bgroup \em et al.\egroup
  }{2022}]{cui2022dc}
Justin Cui, Ruochen Wang, Si~Si, and Cho-Jui Hsieh.
\newblock Dc-bench: Dataset condensation benchmark.
\newblock {\em arXiv preprint arXiv:2207.09639}, 2022.

\bibitem[\protect\citeauthoryear{Deng \bgroup \em et al.\egroup
  }{2009}]{deng2009imagenet}
Jia Deng, Wei Dong, Richard Socher, Li-Jia Li, Kai Li, and Li~Fei-Fei.
\newblock Imagenet: A large-scale hierarchical image database.
\newblock In {\em 2009 IEEE conference on computer vision and pattern
  recognition}, pages 248--255. Ieee, 2009.

\bibitem[\protect\citeauthoryear{Deng}{2022}]{deng2022remember}
Olga Deng, Zhiwei;~Russakovsky.
\newblock Remember the past: Distilling datasets into addressable memories for
  neural networks.
\newblock {\em arXiv preprint arXiv:2206.02916}, 2022.

\bibitem[\protect\citeauthoryear{Dong \bgroup \em et al.\egroup
  }{2022}]{dong2022privacy}
Tian Dong, Bo~Zhao, and Lingjuan Lyu.
\newblock Privacy for free: How does dataset condensation help privacy?
\newblock {\em arXiv preprint arXiv:2206.00240}, 2022.

\bibitem[\protect\citeauthoryear{Dwork}{2006}]{dwork2006differential}
Cynthia Dwork.
\newblock Differential privacy.
\newblock In {\em Automata, Languages and Programming: 33rd International
  Colloquium, ICALP 2006, Venice, Italy, July 10-14, 2006, Proceedings, Part II
  33}, pages 1--12. Springer, 2006.

\bibitem[\protect\citeauthoryear{Fredrikson \bgroup \em et al.\egroup
  }{2015}]{fredrikson2015model}
Matt Fredrikson, Somesh Jha, and Thomas Ristenpart.
\newblock Model inversion attacks that exploit confidence information and basic
  countermeasures.
\newblock In {\em Proceedings of the 22nd ACM SIGSAC conference on computer and
  communications security}, pages 1322--1333, 2015.

\bibitem[\protect\citeauthoryear{Geng \bgroup \em et al.\egroup
  }{2021}]{geng2021towards}
Jiahui Geng, Yongli Mou, Feifei Li, Qing Li, Oya Beyan, Stefan Decker, and
  Chunming Rong.
\newblock Towards general deep leakage in federated learning.
\newblock {\em arXiv preprint arXiv:2110.09074}, 2021.

\bibitem[\protect\citeauthoryear{Geng \bgroup \em et al.\egroup
  }{2023}]{geng2023improved}
Jiahui Geng, Yongli Mou, Qing Li, Feifei Li, Oya Beyan, Stefan Decker, and
  Chunming Rong.
\newblock Improved gradient inversion attacks and defenses in federated
  learning.
\newblock {\em IEEE Transactions on Big Data}, 2023.

\bibitem[\protect\citeauthoryear{Graves \bgroup \em et al.\egroup
  }{2017}]{graves2017automated}
Alex Graves, Marc~G Bellemare, Jacob Menick, Remi Munos, and Koray Kavukcuoglu.
\newblock Automated curriculum learning for neural networks.
\newblock In {\em international conference on machine learning}, pages
  1311--1320. PMLR, 2017.

\bibitem[\protect\citeauthoryear{Hu \bgroup \em et al.\egroup
  }{2022}]{hu2022fedsynth}
Shengyuan Hu, Jack Goetz, Kshitiz Malik, Hongyuan Zhan, Zhe Liu, and Yue Liu.
\newblock Fedsynth: Gradient compression via synthetic data in federated
  learning.
\newblock {\em arXiv preprint arXiv:2204.01273}, 2022.

\bibitem[\protect\citeauthoryear{Jacot \bgroup \em et al.\egroup
  }{2018}]{jacot2018neural}
Arthur Jacot, Franck Gabriel, and Cl{\'e}ment Hongler.
\newblock Neural tangent kernel: Convergence and generalization in neural
  networks.
\newblock {\em Advances in neural information processing systems}, 31, 2018.

\bibitem[\protect\citeauthoryear{Jin \bgroup \em et al.\egroup
  }{2021}]{jin2021graph}
Wei Jin, Lingxiao Zhao, Shichang Zhang, Yozen Liu, Jiliang Tang, and Neil Shah.
\newblock Graph condensation for graph neural networks.
\newblock {\em arXiv preprint arXiv:2110.07580}, 2021.

\bibitem[\protect\citeauthoryear{Jin \bgroup \em et al.\egroup
  }{2022}]{jin2022condensing}
Wei Jin, Xianfeng Tang, Haoming Jiang, Zheng Li, Danqing Zhang, Jiliang Tang,
  and Bing Yin.
\newblock Condensing graphs via one-step gradient matching.
\newblock In {\em Proceedings of the 28th ACM SIGKDD Conference on Knowledge
  Discovery and Data Mining}, pages 720--730, 2022.

\bibitem[\protect\citeauthoryear{Kim \bgroup \em et al.\egroup
  }{2022}]{kim2022dataset}
Jang-Hyun Kim, Jinuk Kim, Seong~Joon Oh, Sangdoo Yun, Hwanjun Song, Joonhyun
  Jeong, Jung-Woo Ha, and Hyun~Oh Song.
\newblock Dataset condensation via efficient synthetic-data parameterization.
\newblock {\em arXiv preprint arXiv:2205.14959}, 2022.

\bibitem[\protect\citeauthoryear{Koehn}{2005}]{koehn-2005-europarl}
Philipp Koehn.
\newblock {E}uroparl: A parallel corpus for statistical machine translation.
\newblock In {\em Proceedings of Machine Translation Summit X: Papers}, pages
  79--86, Phuket, Thailand, September 13-15 2005.

\bibitem[\protect\citeauthoryear{Konyushkova \bgroup \em et al.\egroup
  }{2017}]{konyushkova2017learning}
Ksenia Konyushkova, Raphael Sznitman, and Pascal Fua.
\newblock Learning active learning from data.
\newblock {\em Advances in neural information processing systems}, 30, 2017.

\bibitem[\protect\citeauthoryear{Lee \bgroup \em et al.\egroup
  }{2022}]{lee2022dataset}
Hae~Beom Lee, Dong~Bok Lee, and Sung~Ju Hwang.
\newblock Dataset condensation with latent space knowledge factorization and
  sharing.
\newblock {\em arXiv preprint arXiv:2208.10494}, 2022.

\bibitem[\protect\citeauthoryear{Li \bgroup \em et al.\egroup
  }{2022}]{li2022dataset}
Guang Li, Ren Togo, Takahiro Ogawa, and Miki Haseyama.
\newblock Dataset distillation for medical dataset sharing.
\newblock {\em arXiv preprint arXiv:2209.14603}, 2022.

\bibitem[\protect\citeauthoryear{Liu \bgroup \em et al.\egroup
  }{2022a}]{liu2022graph}
Mengyang Liu, Shanchuan Li, Xinshi Chen, and Le~Song.
\newblock Graph condensation via receptive field distribution matching.
\newblock {\em arXiv preprint arXiv:2206.13697}, 2022.

\bibitem[\protect\citeauthoryear{Liu \bgroup \em et al.\egroup
  }{2022b}]{liu2022dataset}
Songhua Liu, Kai Wang, Xingyi Yang, Jingwen Ye, and Xinchao Wang.
\newblock Dataset distillation via factorization.
\newblock {\em arXiv preprint arXiv:2210.16774}, 2022.

\bibitem[\protect\citeauthoryear{Liu \bgroup \em et al.\egroup
  }{2023}]{liu2023backdoor}
Yugeng Liu, Zheng Li, Michael Backes, Yun Shen, and Yang Zhang.
\newblock Backdoor attacks against dataset distillation.
\newblock {\em arXiv preprint arXiv:2301.01197}, 2023.

\bibitem[\protect\citeauthoryear{Loo \bgroup \em et al.\egroup
  }{2022}]{loo2022efficient}
Noel Loo, Ramin Hasani, Alexander Amini, and Daniela Rus.
\newblock Efficient dataset distillation using random feature approximation.
\newblock {\em arXiv preprint arXiv:2210.12067}, 2022.

\bibitem[\protect\citeauthoryear{Maclaurin \bgroup \em et al.\egroup
  }{2015}]{maclaurin2015gradient}
Dougal Maclaurin, David Duvenaud, and Ryan Adams.
\newblock Gradient-based hyperparameter optimization through reversible
  learning.
\newblock In {\em International conference on machine learning}, pages
  2113--2122. PMLR, 2015.

\bibitem[\protect\citeauthoryear{Nguyen \bgroup \em et al.\egroup
  }{2020}]{nguyen2020dataset}
Timothy Nguyen, Zhourong Chen, and Jaehoon Lee.
\newblock Dataset meta-learning from kernel ridge-regression.
\newblock {\em arXiv preprint arXiv:2011.00050}, 2020.

\bibitem[\protect\citeauthoryear{Nguyen \bgroup \em et al.\egroup
  }{2021}]{nguyen2021dataset}
Timothy Nguyen, Roman Novak, Lechao Xiao, and Jaehoon Lee.
\newblock Dataset distillation with infinitely wide convolutional networks.
\newblock {\em Advances in Neural Information Processing Systems},
  34:5186--5198, 2021.

\bibitem[\protect\citeauthoryear{Platt and
  others}{1999}]{platt1999probabilistic}
John Platt et~al.
\newblock Probabilistic outputs for support vector machines and comparisons to
  regularized likelihood methods.
\newblock {\em Advances in large margin classifiers}, 10(3):61--74, 1999.

\bibitem[\protect\citeauthoryear{Prabhu \bgroup \em et al.\egroup
  }{2020}]{prabhu2020gdumb}
Ameya Prabhu, Philip~HS Torr, and Puneet~K Dokania.
\newblock Gdumb: A simple approach that questions our progress in continual
  learning.
\newblock In {\em European conference on computer vision}, pages 524--540.
  Springer, 2020.

\bibitem[\protect\citeauthoryear{Schmidhuber}{2015}]{schmidhuber2015deep}
J{\"u}rgen Schmidhuber.
\newblock Deep learning in neural networks: An overview.
\newblock {\em Neural networks}, 61:85--117, 2015.

\bibitem[\protect\citeauthoryear{Sener and Savarese}{2017}]{sener2017active}
Ozan Sener and Silvio Savarese.
\newblock Active learning for convolutional neural networks: A core-set
  approach.
\newblock {\em arXiv preprint arXiv:1708.00489}, 2017.

\bibitem[\protect\citeauthoryear{Shokri \bgroup \em et al.\egroup
  }{2017}]{shokri2017membership}
Reza Shokri, Marco Stronati, Congzheng Song, and Vitaly Shmatikov.
\newblock Membership inference attacks against machine learning models.
\newblock In {\em 2017 IEEE symposium on security and privacy (SP)}, pages
  3--18. IEEE, 2017.

\bibitem[\protect\citeauthoryear{Such \bgroup \em et al.\egroup
  }{2020}]{such2020generative}
Felipe~Petroski Such, Aditya Rawal, Joel Lehman, Kenneth Stanley, and Jeffrey
  Clune.
\newblock Generative teaching networks: Accelerating neural architecture search
  by learning to generate synthetic training data.
\newblock In {\em International Conference on Machine Learning}, pages
  9206--9216. PMLR, 2020.

\bibitem[\protect\citeauthoryear{Sucholutsky and
  Schonlau}{2021}]{sucholutsky2021soft}
Ilia Sucholutsky and Matthias Schonlau.
\newblock Soft-label dataset distillation and text dataset distillation.
\newblock In {\em 2021 International Joint Conference on Neural Networks
  (IJCNN)}, pages 1--8. IEEE, 2021.

\bibitem[\protect\citeauthoryear{Tsilivis \bgroup \em et al.\egroup
  }{2022}]{tsilivis2022can}
Nikolaos Tsilivis, Jingtong Su, and Julia Kempe.
\newblock Can we achieve robustness from data alone?
\newblock {\em arXiv preprint arXiv:2207.11727}, 2022.

\bibitem[\protect\citeauthoryear{Wang \bgroup \em et al.\egroup
  }{2018}]{wang2018dataset}
Tongzhou Wang, Jun-Yan Zhu, Antonio Torralba, and Alexei~A Efros.
\newblock Dataset distillation.
\newblock {\em arXiv preprint arXiv:1811.10959}, 2018.

\bibitem[\protect\citeauthoryear{Wang \bgroup \em et al.\egroup
  }{2022}]{wang2022cafe}
Kai Wang, Bo~Zhao, Xiangyu Peng, Zheng Zhu, Shuo Yang, Shuo Wang, Guan Huang,
  Hakan Bilen, Xinchao Wang, and Yang You.
\newblock Cafe: Learning to condense dataset by aligning features.
\newblock In {\em Proceedings of the IEEE/CVF Conference on Computer Vision and
  Pattern Recognition}, pages 12196--12205, 2022.

\bibitem[\protect\citeauthoryear{Warden}{2018}]{warden2018speech}
Pete Warden.
\newblock Speech commands: A dataset for limited-vocabulary speech recognition.
\newblock {\em arXiv preprint arXiv:1804.03209}, 2018.

\bibitem[\protect\citeauthoryear{Wu \bgroup \em et al.\egroup
  }{2022}]{wu2022towards}
Yihan Wu, Xinda Li, Florian Kerschbaum, Heng Huang, and Hongyang Zhang.
\newblock Towards robust dataset learning.
\newblock {\em arXiv preprint arXiv:2211.10752}, 2022.

\bibitem[\protect\citeauthoryear{Xiong \bgroup \em et al.\egroup
  }{2022}]{xiong2022feddm}
Yuanhao Xiong, Ruochen Wang, Minhao Cheng, Felix Yu, and Cho-Jui Hsieh.
\newblock Feddm: Iterative distribution matching for communication-efficient
  federated learning.
\newblock {\em arXiv preprint arXiv:2207.09653}, 2022.

\bibitem[\protect\citeauthoryear{Zhao and Bilen}{2021}]{Zhao2021DatasetCW}
Bo~Zhao and Hakan Bilen.
\newblock Dataset condensation with distribution matching.
\newblock {\em ArXiv}, abs/2110.04181, 2021.

\bibitem[\protect\citeauthoryear{Zhao and Bilen}{2022}]{zhao2022synthesizing}
Bo~Zhao and Hakan Bilen.
\newblock Synthesizing informative training samples with gan.
\newblock {\em arXiv preprint arXiv:2204.07513}, 2022.

\bibitem[\protect\citeauthoryear{Zhao \bgroup \em et al.\egroup
  }{2021}]{zhao2021datasetdc}
Bo~Zhao, Konda~Reddy Mopuri, and Hakan Bilen.
\newblock Dataset condensation with gradient matching.
\newblock {\em ICLR}, 1(2):3, 2021.

\bibitem[\protect\citeauthoryear{Zhao~Bo}{2021}]{zhao2021datasetdsa}
Bilen~Hakan Zhao~Bo.
\newblock Dataset condensation with differentiable siamese augmentation.
\newblock In {\em International Conference on Machine Learning}, pages
  12674--12685. PMLR, 2021.

\bibitem[\protect\citeauthoryear{Zhou \bgroup \em et al.\egroup
  }{2022}]{zhou2022dataset}
Yongchao Zhou, Ehsan Nezhadarya, and Jimmy Ba.
\newblock Dataset distillation using neural feature regression.
\newblock {\em arXiv preprint arXiv:2206.00719}, 2022.

\end{thebibliography}

\end{document}